\documentclass[letterpaper, 10 pt, conference]{ieeeconf}
\usepackage[final]{microtype}
\IEEEoverridecommandlockouts

\usepackage{afterpage}
\usepackage{svg}
\usepackage{siunitx}
\usepackage{hyperref}
\usepackage{url}
\usepackage{xspace}
\usepackage{amssymb,paralist,epsfig,standalone,bm,placeins}  
\usepackage{graphicx} 
\usepackage{xcolor}
\usepackage{multirow}
\usepackage{makecell}
\usepackage{tabularx}
\usepackage{multicol}
\usepackage[utf8]{inputenc}
\usepackage{amsmath,amsfonts,booktabs,cite} 
\usepackage{siunitx,textcomp} 
\let\labelindent\relax
\usepackage[inline]{enumitem} 
\usepackage[nolist]{acronym} 
\usepackage{caption}
\usepackage{tikz,standalone,pgfplots}
\pgfplotsset{width=10cm,compat=1.9}
\usetikzlibrary{patterns}
\usepackage{pgfplots, pgfplotstable}
\graphicspath{{figures_tex/}}
\usepackage{changes}
\usepackage[ruled,vlined]{algorithm2e}
\usepackage{subcaption}

\makeatletter
\newcommand{\removelatexerror}{\let\@latex@error\@gobble}
\let\@oldmaketitle\@maketitle
\renewcommand{\@maketitle}{\@oldmaketitle
}
\makeatother
\pgfplotsset{compat=1.3}
\newacro{ftsensor}[F/T sensor]{force-torque sensor}
\newacro{ecoc}[ECOC]{error-correcting output codes}
\newacro{svm}[SVM]{support vector machines}

\newacro{pbd}[PBD]{Position-based Dynamics}
\newacro{fem}[FEM]{Finite Element Method}
\newacro{dnn}[DNN]{Deep Neural Network}
\newacro{fcn}[FCN]{fully-convolutional network}

\newcommand{\figref}[1]{\hyperref[#1]{Fig.~\ref*{#1}}}
\newcommand{\tabref}[1]{\hyperref[#1]{Table~\ref*{#1}}}
\newcommand{\secref}[1]{\hyperref[#1]{Section~\ref*{#1}}}
\newcommand{\algoref}[1]{\hyperref[#1]{Algorithm~\ref*{#1}}}

\definecolor{findOptimalPartition}{HTML}{D7191C}
\definecolor{storeClusterComponent}{HTML}{FDAE61}
\definecolor{dbscan}{HTML}{ABDDA4}
\definecolor{constructCluster}{HTML}{2B83BA}

\title{\LARGE \bf Model-Predictive Spill-Free Emergency Stopping\\ for Liquid-Manipulating Robots}

\author{Samuli Hynninen$^{1}$, Ville Kyrki $^{1}$%
\thanks{This work was financially supported by the Finnish Ministry of Education and Culture through the Intelligent Work Machines Doctoral Education Pilot Program (IWM VN/3137/2024-OKM-4).}
\thanks{$^{1}$ Intelligent Robotics Group at the Department of Electrical Engineering and Automation, School of Electrical Engineering, Aalto University, Finland. \texttt{\{firstname.lastname\}{@}aalto.fi}}
}

\begin{document}
\maketitle
\thispagestyle{empty}
\pagestyle{empty}


\begin{abstract}
Manipulating open liquid containers is challenging because liquids are highly sensitive to vessel accelerations and jerks. Although spill-free liquid manipulation has been widely studied, emergency stopping under unexpected hazards has received little attention, despite the fact that abrupt braking may cause hazardous spills. This letter presents an emergency stop system for robots manipulating liquids in open containers. We formulate emergency stopping as an optimal control problem and solve it in a model predictive control framework to generate time-optimal, spill-free stopping trajectories. The method operates as a plug-and-play safety layer on top of existing slosh-free motion planning methods, enabling immediate reaction to detected hazards while accounting for nonlinear liquid dynamics. We demonstrate, through simulation and on a 7-DoF Franka Emika Panda robot, that the proposed approach achieves fast emergency stopping without spilling.

\end{abstract}

\section{Introduction}
\label{sec:introduction}
The ability to manipulate open liquid containers is vital in various applications of robotics, for example, when assistant robots serve beverages or industrial robots transfer chemicals in factory settings. However, manipulating such containers is notoriously difficult because liquids exhibit complex nonlinear dynamics and react easily to even small accelerations of the container. 

For that reason, slosh suppression, as well as spill- and slosh-free manipulation, are widely studied problems in robotic manipulation  \cite{abderezaei2024clutter,aribowo2015integrated,arrizabalaga2024geometric,di2021time,maderna2018robotic,moriello2017control,muchacho2022solution,reinhold2019dynamic, biagiotti2018plug, zang2014dynamics, pridgen2010slosh, chen2007using, consolini2013minimum}. Proposed solutions mostly rely on simplified mechanical slosh models to avoid computationally heavy fluid dynamics simulations, or omit modeling and focus on minimizing the slosh-inducing forces acting on the container.

Despite extensive research, none of the existing methods account for unexpected execution-time collision hazards, such as those caused by unpredictable human behavior, that necessitate rapid stopping of the robot for safety. Even if the collision can be avoided by a prompt reaction, the liquid inside the container (e.g., hot coffee) may cause severe damage if spilled due to overly aggressive emergency braking. For this reason, stopping must be performed both quickly and safely. Such a stopping strategy requires careful consideration of the trade-off between rapid deceleration and induced slosh. Furthermore, the problem is particularly challenging because the complex, nonlinear liquid dynamics must be incorporated into a real-time model that provides optimal stopping trajectories for arbitrary initial states without delay. 

To this end, we propose an emergency stopping method that enables immediate reaction and brings a liquid-carrying robot to rest in a time-optimal manner while ensuring that no liquid is spilled. We achieve this by formulating the emergency stop as an optimal control problem and solving it repeatedly in a model predictive control (MPC) fashion. The proposed method works as a plug-and-play safety feature on top of existing slosh-free motion planning methods, such as \cite{arrizabalaga2024geometric, muchacho2022solution}. Furthermore, we carefully address the nonlinearities of our liquid dynamics model to enable fast computations without compromising the model's predictive power or limiting the robot's motion directions in 6D space. The proposed method is robust to moderate modeling errors, and the liquid model used can be applied to various container sizes and shapes \cite{ibrahim2005liquid, ibrahim2001recent}.

To summarize, this work makes the following contributions
\begin{itemize}
    \item We formalize the problem of emergency stopping in liquid manipulation. The problem has applications across various domains, from household robotics to industrial manufacturing.
    \item We propose an MPC-style emergency stopping approach that achieves rapid stopping while preventing liquid spillage. The approach serves as a plug-and-play safety layer atop existing slosh-free motion planning methods.
    \item We provide a comprehensive analysis of the proposed method through simulation experiments and demonstrate its performance on real robot hardware using actual water as an example liquid.
\end{itemize}
The rest of the paper first reviews related work in Section \ref{sec:related_work} and then presents the liquid slosh and robot manipulator models in Section \ref{sec:background}. The proposed method and experimental results follow in Sections \ref{sec:method} and \ref{sec:exp_and_res}, respectively, before Section \ref{sec:conclusions} concludes the paper.
\section{Related work}
\label{sec:related_work}
\subsection{Liquid manipulation}
Spill-free liquid manipulation and slosh suppression are extensively studied problems in robotics and automation. Contributions include input shaping methods~\cite{pridgen2010slosh, aribowo2015integrated, moriello2017control, biagiotti2018plug, zang2014dynamics}, which analyze the liquid's natural sloshing frequencies in the given container and use filtering techniques to adjust control inputs that would otherwise excite sloshing. Slosh-modeling-based trajectory optimization methods are introduced in \cite{consolini2013minimum, reinhold2019dynamic, maderna2018robotic, di2021time}. In these methods, a simplified mechanical model of liquid dynamics is used, and the goal is to plan the trajectory from the beginning so that slosh is minimized or remains within given limits while still being time-efficient. An alternative trajectory optimization approach is introduced in \cite{muchacho2022solution}, in which the authors use a concept of virtual pendulum to mimic the behaviour of a mechanical slosh-free tray called SpillNot \cite{millstein2012tray}.  Furthermore, \cite{arrizabalaga2024geometric} introduces a trajectory optimization approach that uses the idea of a virtual quadrotor to ensure that accelerations acting on the container are perpendicular to the liquid surface, and thus, slosh-free. Several works \cite{di2026sloshing, DiLeva2022SloshingDE,guagliumi} focus on slosh modeling and estimation. 
In recent work, \cite{abderezaei2024clutter} utilizes learned dynamics and a transformer-based neural network to generate clutter-aware paths and trajectories without spill, while  \cite{ferrari} considers multi-container manipulation. 

Despite numerous methods for spill-free liquid manipulation, none address hazards that arise during execution and require real-time replanning for safety. We introduce an emergency trajectory planning method that enables an immediate reaction to a detected hazard and brings the system to a stop in a safe, time-optimal manner. Our method falls into a similar category of trajectory optimization to \cite{consolini2013minimum, reinhold2019dynamic, maderna2018robotic, di2021time, muchacho2022solution, arrizabalaga2024geometric}, but unlike these works, we do not have a pre-specified reference path or goal pose to follow. To ensure spill-freeness of the emergency stop trajectory, we use a similar slosh modeling approach to \cite{consolini2013minimum, reinhold2019dynamic, maderna2018robotic, di2021time}.

\subsection{Emergency stop}
Emergency stop has been widely studied in fields such as autonomous driving \cite{svensson2018safe, wang2020real, krook2019design,yang2020control, kwon2014autonomous}. While these works share the same goal, the environment and related assumptions differ substantially from those in our case. 
In robotic manipulation, emergency stopping has received little attention. Prior works \cite{de2006collision, haddadin2008collision} treat it primarily as a collision response, but because they consider empty manipulators without non-prehensile or unactuated components, stopping reduces to trivial full-force braking.

The closest related setting is emergency braking for overhead cranes \cite{chen2019new, ma2010switching, reddy2024emergency, raj2023emergency}, where payload swing resembles liquid slosh. However, these systems are limited by their hardware setting to 1D linear motion and one-directional swinging, whereas our problem involves 6D motion, 2-DoF sloshing, and a multi-DoF manipulator

\section{Background}
\label{sec:background}
\subsection{Problem description and solution overview}
\label{problem_description}
We address the problem of bringing a robotic manipulator arm carrying a liquid container to rest as quickly as possible, without spilling the liquid, after an emergency stop is triggered. Formally, this means minimizing the cumulative sum of the Cartesian velocities while obeying no-spill constraints for the allowed liquid-surface inclination.

Importantly, our emergency stop system builds on existing slosh-free motion generation methods, such as \cite{arrizabalaga2024geometric, muchacho2022solution}. Slosh-free motion planning guarantees that, at the start of the emergency stop, the liquid surface normal is aligned with the container's vertical axis. Thus, we know the liquid's initial state without challenging real-time perception, and we have a starting point for modeling its evolution. 

The proposed method uses a standard model predictive control approach, in which we repeatedly solve optimal control problems with a receding time horizon. For slosh modeling, we use a well-established spherical pendulum model for its computational efficiency and quantitatively demonstrated modeling accuracy \cite{di2026sloshing}.

Even with the simplified fluid dynamics model, the problem remains highly nonlinear and thus requires intricate linearization efforts to enable real-time tractability without sacrificing modeling power. Ultimately, we formulate the problem as a Quadratic Program (QP) and first solve it in the task space. The task-space solution is then converted into the joint-space using the Resolved Acceleration Control (RAC) approach \cite{luh2003resolved}, that is, by solving another small-scale QP. A direct joint-space MPC, on the other hand, would require predicting the configuration-dependent Jacobian and its derivative for every time step in the horizon, resulting in a larger nonlinear program that is less suitable for real-time control. In the proposed dual-stage approach, the joint-space conversion is only made for the first time step, keeping the problem small and linear. 
\subsection{Liquid slosh modelling} 
\subsubsection{Spherical pendulum model}
To model the liquid slosh, we use the well-established spherical pendulum model \cite{ibrahim2005liquid}, whose state is parametrized by two tilt angles ($\theta_p, \phi_p$) and the pendulum rod length ($l$). We adopt the fixed-world-axis rotation sequence $R_x(\phi_p)R_y(\theta_p)$, i.e., a rotation by $\theta_p$ about the world-frame $y$-axis followed by a rotation by $\phi_p$ about the world-frame $x$-axis. Accordingly, the position of the pendulum mass $\mathbf{p}_m = [x_m, y_m, z_m]^\top$, given the pivot position $\mathbf{p} = [x, y, z]^\top$, is given by
\begin{equation}
    \mathbf{p}_m =\begin{bmatrix}
        x - l\sin\theta_p \\
        y + l\cos\theta_p\sin\phi_p \\
        z - l\cos\theta_p\cos\phi_p
    \end{bmatrix}.
\end{equation}
This position and its time derivative can be used to determine the system's potential and kinetic energy. By utilizing the system's energy, we can build Euler-Lagrange equations for the system, and as a result, we can write the system's equations of motion as
\begin{equation}
    \begin{split}
    \label{nonlinear-dynamics-1}
    l\ddot{\theta}_p = \ &(g+\ddot{z})\,\sin\theta_p\,\cos\phi_p
+\ddot{x}\cos\theta_p
\\ &+\ddot{y}\,\sin\phi_p\,\sin\theta_p 
-l\,\cos\theta_p\,\sin\theta_p\,\dot{\phi}^{2}_p
 \end{split}
\end{equation}
\begin{equation}
\label{nonlinear-dynamics-2}
l\cos\theta_p\,\ddot{\phi}_p =
(g+\ddot{z})\,\sin\phi_p
-\ddot{y}\cos\phi_p
+2l\,\dot{\phi}_p\,\dot{\theta}_p\,\sin\theta_p
\end{equation}
where $\ddot{x}, \ddot{y}, \ddot{z}$ refer to pendulum pivot point accelerations.

Using the full nonlinear equations in the optimal control problem would make real-time solution infeasible, and therefore, we linearize the equations.  Following the traditional convention (see e.g. \cite{maderna2018robotic, biagiotti2018plug, reinhold2019dynamic}) in the slosh-free manipulation literature, we assume small angles and linearize the dynamics about a stable equilibrium at $\theta_p = \phi_p = \dot{\theta}_p = \dot{\phi}_p = 0$. The higher-order terms in~(\ref{nonlinear-dynamics-1}),~(\ref{nonlinear-dynamics-2}) become products of small perturbations and can therefore be neglected in the first-order linearized model.   
The linearized small-angle dynamics
\begin{equation}
    \ddot{\theta}_p = \frac{g + \ddot{z}}{l}\theta_p + \frac{1}{l}\ddot{x}
\end{equation}
\begin{equation}
    \ddot{\phi}_p = \frac{g + \ddot{z}}{l}\phi_p - \frac{1}{l}\ddot{y}.
\end{equation}
still contain problematic bilinear relationships between $\ddot{z}$ and $\theta_p$ and $\phi_p$, which cannot be handled by the quadratic program. A common solution (e.g., in \cite{maderna2018robotic, biagiotti2018plug, reinhold2019dynamic}), is to consider also $\ddot{z}$ as a small perturbation and neglect the products $\ddot{z}\theta_p$, $\ddot{z}\phi_p$. However, this removes the coupling between vertical acceleration and the pendulum state, effectively restricting the model to planar motion. We consider this an unacceptable oversimplification for an emergency stop system.

For this reason, we linearize
the problematic bilinear multiplications around a nominal trajectory (solution from the previous round) for each time step in the optimization horizon. Formally, we approximate the product $\ddot{z}\theta_{p}$ as
\begin{equation}
    \ddot{z}\theta_{p} \approx \bar{\ddot{z}}\theta_{p} + \bar{\theta}_{p}\ddot{z} - \bar{\ddot{z}} \bar{\theta}_{p},
\end{equation}
where the nominal values $\bar{\ddot{z}}$ and $ \bar{\theta}_{p}$ are obtained from the previous solution. The same approximation is applied for the product $\ddot{z}\phi_{p}$.

\subsubsection{Sloshing height}
\label{sec:slosh_height}
The spherical pendulum model assumes that the liquid free surface remains perpendicular to the pendulum rod. The so-called sloshing angle, $\beta$, is defined as the angle between the pendulum rod and the container's vertical axis. If the container tilt angles are denoted by $\theta_c, \phi_c$, the sloshing angle is given as 
\begin{equation}
\label{eq:sloshing_angle}
\beta = \cos^{-1}(\mathbf{n}_p\cdot\mathbf{n}_c),   
\end{equation}
where
\begin{equation}
    \mathbf{n}_p = \begin{bmatrix}
        -\sin\theta_p \\
        \cos\theta_p\sin\phi_p\\
        -\cos\theta_p\cos\phi_p
    \end{bmatrix} \ \text{and} \ \mathbf{n}_c = \begin{bmatrix}
        -\sin\theta_c \\
        \cos\theta_c\sin\phi_c\\
        -\cos\theta_c\cos\phi_c
    \end{bmatrix}
\end{equation}
are unit vectors aligned with the pendulum rod and the container's vertical axis, respectively. For a cylindrical container, the sloshing height $\bar{\eta}$ is given as $\bar{\eta} = R\tan\beta$, where $R$ is the container radius. Conversely, the sloshing angle can be obtained from the sloshing height as $\beta = \tan^{-1}(\bar{\eta}/R)$.

\subsubsection{Rod length estimation}
\label{rod_length_estimation}
To use the pendulum model, the rod length must be estimated based on the container's shape and size. A well-established \cite{ibrahim2001recent, ibrahim2005liquid} method to determine the rod length $l$ uses the natural frequencies of slosh. For a cylindrical container, the natural frequency of the first sloshing mode is
\begin{equation}
    f_n = \sqrt{ \frac{g\xi}{R} \tanh \Big( \frac{h\xi}{R}}\Big),
\end{equation}
where $\xi$ is the first root of the derivative of the Bessel function of the first kind, $R$ is the radius of the container, and $h$ is the liquid height. This is related to the rod length by an equality $f_n = \sqrt{g/l}$,
which we can solve for $l$. While the upward acceleration $\ddot{z}$ theoretically affects the effective gravity and makes the rod length time-varying, we opt for a constant rod length, assuming $\ddot{z} = 0$, for simplicity. This approach can be easily extended to non-cylindrical containers, but we refer the reader to \cite{ibrahim2005liquid} for further details. 

\subsection{Robot manipulator model}
The relation between the robot's Cartesian pose, $\mathbf{X} \in SE(3)$, and the robot joint angles, $\mathbf{q} \in \mathbb{R}^n$, is determined by forward kinematics
\begin{equation}
    \mathbf{X} = f(\mathbf{q}), 
\qquad
f:\mathbb{R}^n \rightarrow SE(3),
\end{equation}
where $f$ can be determined from Denavit-Hartenberg parameters \cite{denavit1955kinematic}  or elementary transform sequences \cite{corke2007simple}.

For the Cartesian twist $\boldsymbol{\xi} \in \mathbb{R}^6$, the relation is derived through the robot's Jacobian $J(\mathbf{q}) \in \mathbb{R}^{6 \times n}$, and can be written as
\begin{equation}
    \label{cart_v_joint_v}
    \boldsymbol{\xi} =  J(\mathbf{q}) \dot{\mathbf{q}}.
\end{equation}
By differentiating Eq. (\ref{cart_v_joint_v}), we obtain
\begin{equation}
    \dot{\boldsymbol{\xi}} = \dot{J}(\mathbf{q}) \dot{\mathbf{q}} + J(\mathbf{q})\ddot{\mathbf{q}},
\end{equation}
which gives the Cartesian acceleration.
\section{Methodology}
\begin{figure*}[h]
    \centering
    \includegraphics[width=0.7\linewidth]{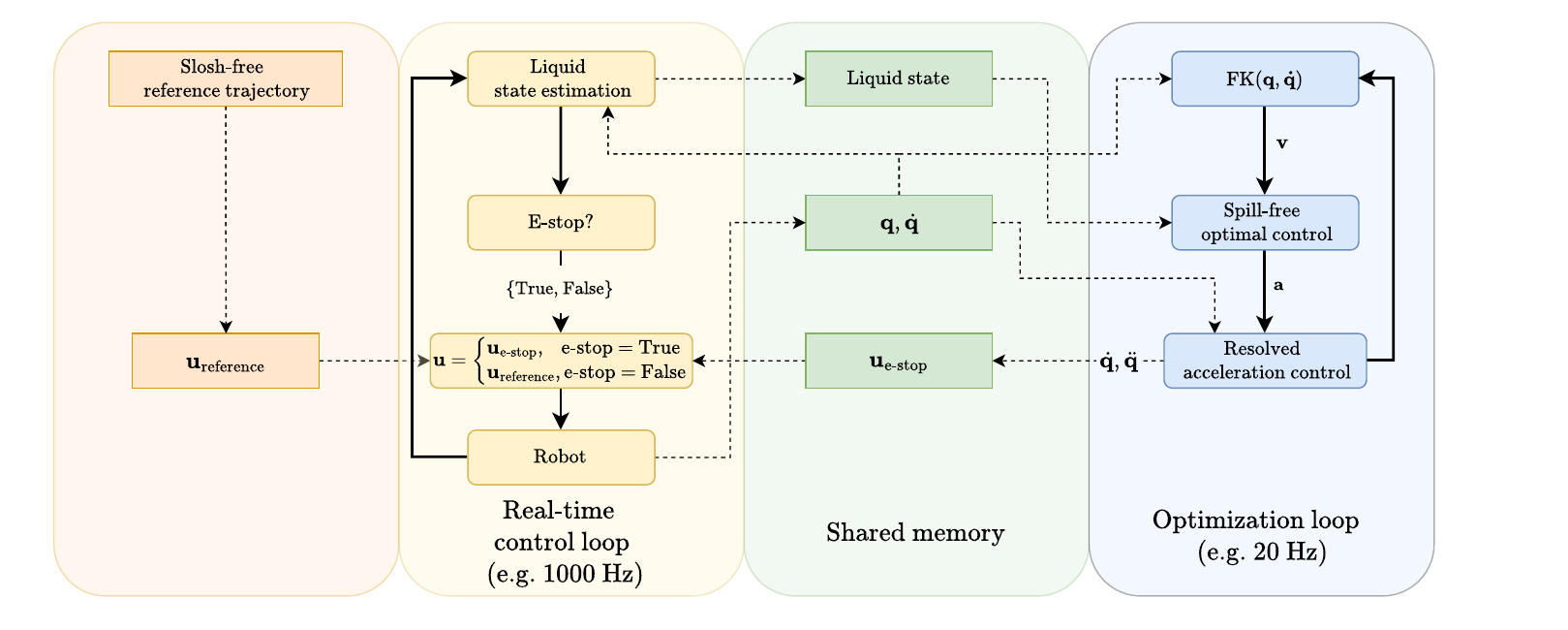}
    \caption{Overview of the proposed emergency stop architecture. The real-time control loop (yellow) communicates with the robot and updates the liquid state estimate, while the optimization loop (blue) continuously computes time-optimal, spill-free stopping trajectories. Green boxes denote shared-memory variables. Before emergency-stop activation, the liquid surface estimate is initialized such that the surface normal is perpendicular to the container’s vertical axis.}
    \label{fig:system_architecture}
\end{figure*}
\label{sec:method}
\subsection{State-space model}
The state space of our model comprises the pendulum angles and angular velocities, the end-effector's 6-DoF Cartesian velocities, and the end-effector's 3-DoF Cartesian orientation, totaling 13 state variables. We can write $\mathbf{s} = [\dot{x}, \dot{y}, \dot{z}, \theta_p, \phi_p, \dot{\theta}_p, \dot{\phi}_p, \theta_c, \phi_c, \psi_c, \dot{\theta}_c, \dot{\phi_c}, \dot{\psi_c}]$, where $\theta_p, \phi_p$ and their derivatives refer to pendulum angles, and $\theta_c, \phi_c, \psi_c$ and their derivatives refer to those of the container. We have six control variables, namely the container's Cartesian accelerations, $\mathbf{u} = [\ddot{x}, \ddot{y}, \ddot{z}, \ddot{\theta_c}, \ddot{\phi_c}, \ddot{\psi_c}]$. 

We have a discretized time-varying state-equation
\begin{equation}
    \mathbf{s}_{k+1} = \mathbf{A}_k\mathbf{s}_k + \mathbf{B}_k \mathbf{u}_k + \mathbf{d}_k,
\end{equation}
where the state transition matrix $\mathbf{A}_k \in \mathbb{R}^{13 \times 13}$ is defined

\begin{equation}
\mathbf{A}_k =
\begin{bmatrix}
\mathbf{I}_3 & 0 & 0 \\
0 & \mathbf{A}_{1,k} & 0 \\
0 & 0 & \mathbf{A}_2
\end{bmatrix},
\end{equation}
with
\begin{equation}
\mathbf{A}_{1,k} =
\begin{bmatrix}
\alpha_k \mathbf{I}_2 & dt\, \mathbf{I}_2 \\
\beta_k \mathbf{I}_2 & \mathbf{I}_2
\end{bmatrix}, \quad
A_2 =
\begin{bmatrix}
    \mathbf{I}_3 & \Delta_t \mathbf{I}_3 \\
    \mathbf{0}_{3 \times 3} & \mathbf{I}_3
\end{bmatrix},
\end{equation}
where
\begin{equation}
    \alpha_k = 1 + (g +\bar{\ddot{z_k}})\frac{\Delta_t^2 }{2l} 
\end{equation}
\begin{equation}
    \beta_k = (g + \bar{\ddot{z_k}})\frac{\Delta_t}{l},
\end{equation}
where $\bar{\ddot{z_k}}$ is the nominal vertical acceleration, around which we linearize. The control matrix $\mathbf{B}_k \in \mathbb{R}^{13 \times 6}$ is defined
\begin{equation}
    \mathbf{B}_k = \begin{bmatrix}
        \mathbf{B}_{1,k} & \mathbf{0}_{7 \times 3} \\
        \mathbf{0}_{6 \times 3} & \mathbf{B}_{2}
    \end{bmatrix}
\end{equation}
with
\begin{align}
    \begin{split}
    \mathbf{B}_{1,k} &=
    \begin{bmatrix}
        \Delta_t \mathbf{I}_3 \\
        \begin{bmatrix}
           \gamma & 0 & \gamma\bar{\theta}_{p,k}\\
0 & -\gamma & \gamma\bar{\phi}_{p,k}\\
\Delta_t/l & 0 & \frac{\Delta_t}{l}\bar{\theta}_{p,k}\\
0 & -\Delta_t/l & \frac{\Delta_t}{l}\bar{\phi}_{p,k} 
        \end{bmatrix}
    \end{bmatrix},
\quad
\mathbf{B}_2 = \begin{bmatrix}
    \frac{\Delta_t^2}{2} \mathbf{I}_3 \\
    \Delta_t \mathbf{I}_3
\end{bmatrix}, \\
\gamma &= \frac{\Delta_t^2}{2l},
\end{split}
\end{align}
where $\bar{\theta}_{p,k}, \bar{\phi}_{p,k}$ are the nominal angles for linearization. Finally, the affine term $\mathbf{d_k}$ is defined as
\begin{equation}
    \mathbf{d}_k = \begin{bmatrix}
        \mathbf{0}_{3\times1} \\
        -\gamma\bar{\ddot{z}}_k\bar{\theta}_{p,k}\\
        -\gamma\bar{\ddot{z}}_k\bar{\phi}_{p,k}\\
        -\frac{\Delta_t}{l}\bar{\ddot{z}}_k\bar{\theta}_{p,k}\\
        -\frac{\Delta_t}{l}\bar{\ddot{z}}_k\bar{\phi}_{p,k}\\
        \mathbf{0}_{6\times1}    \end{bmatrix}
\end{equation}

\subsection{QP formulation}
\label{sec:qp_formulation}
We formulate the optimal control problem as a quadratic program. Our objective is to minimize the cumulative sum of the container's Cartesian velocities, $\mathbf{v}_k$, while respecting the robot's kinematic limits and the no-spill constraints. For a rigidly held container $ \mathbf{v}_k = \boldsymbol{\xi}_k$, so the objective is equivalent to minimizing the end-effector's velocities. The no-spill constraints are introduced as soft constraints, with a high cost on the non-negative slack variables $\delta_{\theta,k}, \delta_{\phi,k}$. Additionally, we add a small jerk cost to smooth the solution space.Here, we define the jerk, $\mathbf{j}_k$, simply as the difference between successive control inputs $\mathbf{j}_k = \mathbf{u}_k - \mathbf{u}_{k-1}$. We emphasize that the jerk cost alone is insufficient for planning spill-free stop trajectories, as we will show in the experimental results. Formally, we can write the problem as
\begin{align}
    \label{eq:cost_function}
    \min_u &\sum_{k=1}^N c_1||\mathbf{v}_k||^2 + c_2||\mathbf{j}_k||^2 + c_3\delta_{\theta,k} + c_4 \delta_{\phi,k}\\
    &\text{s.t.} \quad \mathbf{s}_{k+1} = \mathbf{A}_k \mathbf{s}_k + \mathbf{B}_k\mathbf{u}_k + \mathbf{d}_k \\
        \label{slosh1}
        &\theta_{\text{min}} \leq \theta_{p,k} - \theta_{c,k} - \delta_{\theta,k}\leq \theta_{\text{max}} \\
    \label{slosh2}
    &\phi_{\text{min}} \leq \phi_{p,k} - \phi_{c,k} - \delta_{\phi,k} \leq \phi_{\text{max}}\\
    \label{kinematic_constraint_1}
    &\mathbf{v}_{\text{min}} \leq \mathbf{v}_k \leq \mathbf{v}_{\text{max}}\\
        \label{kinematic_constraint_2}
        &\mathbf{a}_{\text{min}} \leq \mathbf{u}_k \leq \mathbf{a}_{\text{max}} \\
        &\delta_{\theta,k},\delta_{\phi,k}\geq 0
\end{align}
where (\ref{slosh1})--(\ref{slosh2}) are the no-spill constraints and (\ref{kinematic_constraint_1})--(\ref{kinematic_constraint_2}) are the robot's velocity and acceleration constraints in Cartesian space, respectively. The no-spill constraints (\ref{slosh1}) and (\ref{slosh2}) are linear approximations of the nonlinear no-spill constraint $\beta \leq \beta_{\max}$, where $\beta$ is the sloshing angle defined in (\ref{eq:sloshing_angle}). We set the bounds $\theta_{\text{max}}, \theta_{\text{min}}, \phi_{\text{max}}, \phi_{\text{min}}$ so that the approximations become conservative. Specifically, we set $\theta_{\max} = \phi_{\max} = \beta_{\max}/\sqrt{2}$ (since $\beta^2 \le (\theta_p-\theta_c)^2 + (\phi_p-\phi_c)^2$) and, by symmetry, $\theta_{\min} = -\theta_{\max} = \phi_{\min} = -\phi_{\max}$. The allowed sloshing angle $\beta_{\max}$ is determined by the available free space between the liquid surface at rest and the container rim, and can be computed as described in \ref{sec:slosh_height}. For constants $c_1, c_2, c_3, c_4$, we set $c_1 = 1$, $c_2=0.1$, $c_3 = c_4 = 1000$. For the combination of a very small rod length ($l \approx 0.01$) and a high initial velocity (linear speed $\gtrsim$ 1 m/s), we recommend also penalizing $\dot{\theta}_{p,k}$ and $\dot{\phi}_{p,k}$ in (\ref{eq:cost_function}) to avoid oscillatory behavior caused by the highly sensitive pendulum model.

\subsection{Resolved Acceleration Control}
To convert the task space control actions into the robot's joint space while respecting the joint-wise kinematic limits, we rely on Resolved Acceleration Control (RAC) \cite{luh2003resolved} with slack variables $\boldsymbol{\delta}$, similarly to \cite{arrizabalaga2024geometric}. This formulation yields another linear quadratic program, which can be solved very fast due to its convexity and small size. The program takes the current joint position and velocity, $\mathbf{q}_0$ and $\mathbf{\dot{q}_0}$, the corresponding Jacobian $J(\mathbf{q}_0)$, its time derivative $\dot{J}(\mathbf{q}_0)$, and the task-space control action $\mathbf{u}$ as inputs. It outputs the desired joint position, velocity, and acceleration $\mathbf{q}, \dot{\mathbf{q}}, \ddot{\mathbf{q}}$. Since the RAC program is solved only for the first control step after each task-space optimization, no time indexing is needed. Formally, the optimization problem can be written
\begin{align}
    \min_{\mathbf{q}, \mathbf{\dot{q}}, \mathbf{\ddot{q}}, \boldsymbol{\delta}}
        &[\mathbf{q}, \mathbf{\dot{q}}, \mathbf{\ddot{q}}, \boldsymbol{\delta} ] \mathbf{P} [\mathbf{q}, \mathbf{\dot{q}}, \mathbf{\ddot{q}}, \boldsymbol{\delta} ]^\top \\
        &\text{s.t.} \quad \mathbf{q} = \mathbf{q}_0 + \mathbf{\dot{q}}\Delta_t \\
        &\mathbf{\dot{q}} = \dot{\mathbf{q}}_0 + \mathbf{\ddot{q}} \Delta_t \\
        &\mathbf{u} + \boldsymbol{\delta} = \dot{J}(\mathbf{q_0})\mathbf{\dot{q}} + J(\mathbf{q}_0)\ddot{\mathbf{q}} \\
        \label{eq:joint_space_kin1}
        &\mathbf{q}_{\text{min}} \leq \mathbf{q} \leq \mathbf{q}_{\text{max}}
        \\
        \label{eq:joint_space_kin2}
        &\mathbf{\dot{q}}_{\text{min}} \leq \mathbf{\dot{q}} \leq \mathbf{\dot{q}}_{\text{max}} \\
        \label{eq:joint_space_kin3}
        &\mathbf{\ddot{q}}_{\text{min}} \leq \mathbf{\ddot{q}} \leq \mathbf{\ddot{q}}_{\text{max}},
\end{align}
where $\mathbf{P} \in \mathbb{R}^{27 \times 27}$ is a diagonal cost matrix defining the weighting between variables. The equations (\ref{eq:joint_space_kin1})--(\ref{eq:joint_space_kin3}) represent the robot's kinematic constraints in joint space. We set $\mathbf{P}=\operatorname{blkdiag}(10^{-8}\mathbf{I}_7,\mathbf{I}_7,10^{-8}\mathbf{I}_7,10^3\mathbf{I}_6)$.
\subsection{Implementation}

We run the optimization in receding horizon fashion at 20 Hz frequency, with $\Delta_t = 1/20$, while a high-frequency real-time control loop (e.g., 1000 Hz for a Franka Emika Panda robot) reads the solution from shared memory and sends the first control action to the robot, as in standard model predictive control. The high-frequency loop also reads the robot state, estimates the pendulum state after the emergency stop has been triggered (by integrating the pendulum dynamics), and writes this information to shared memory, where it serves as input to the optimization. The system architecture is also visualized in Figure \ref{fig:system_architecture}. The proposed architecture enables an immediate reaction to an emergency stop trigger while continuously reoptimizing the stopping plan.  At each optimization iteration, we use the previous solution as the nominal trajectory for linearizing the bilinear terms in the system dynamics. The optimization horizon is 2 seconds, corresponding to 40 time steps.  


\section{{Experiments}}
\label{sec:exp_and_res}
\subsection{Simulation analysis}
\label{sec:simulation}
The simulation analysis aimed to evaluate how stopping time depends on the allowed slosh and pendulum rod length, how robust the model is to modeling errors and deviations from the zero initial sloshing angle assumption, and how well it generalizes across diverse initial conditions. We performed the analysis in a custom-made simulator. The simulator models the liquid as a spherical pendulum using the full nonlinear dynamics. The simulator's update frequency was set to 60 Hz. We did not include a robotic arm for simplicity. Thus, the simulation analysis was limited to the Cartesian task space. To match the later real-world experiments, the kinematic constraints in (\ref{kinematic_constraint_1})--(\ref{kinematic_constraint_2}) were set as conservative linear approximations of the Cartesian kinematic limits of the Franka Emika Panda robot. For each translational or rotational magnitude limit $\ell$ given in the hardware specification \cite{franka_robot_specs}, we imposed the per-axis bounds [$-\ell / \sqrt{3}, \ell / \sqrt{3}$]. This ensures that any solution satisfying the per-axis bounds also satisfies the corresponding Cartesian magnitude limit.

In the first simulation experiment, we analyzed how rod length and allowed maximum slosh ($\beta_{\max}$), determined by the container size and geometry, affect the stopping time. As the test trajectory, the container was accelerated uniformly along each axis to a speed of 1 m/s over a 1-second acceleration period, with the acceleration ramped smoothly to smoothen the pendulum behaviour. Stopping was considered complete when the container speed fell below $10^{-2}$ m/s. This tolerance was used because small-amplitude oscillations around zero often remain after deceleration, and the threshold better captures the practically relevant part of the speed curve. 

\begin{figure}[!t]
\centering
\begin{tikzpicture}
\begin{axis}[
    width=0.7\columnwidth,
    height=0.7\columnwidth,
    xlabel={$\beta_{\max} $ ($^\circ$)},
    ylabel={Rod length (m)},
    colormap/viridis,
    colorbar,
    colorbar style={ylabel={Stopping time (s)}},
    xmin=-0.5, xmax=14.5,
    ymin=-0.5, ymax=13.5,
    y dir=reverse,
    enlargelimits=false,
    point meta min=0.9,
    point meta max=1.4,
    xtick={0,2,4,6,8,10,12,14},
    xticklabels={1, 3, 5, 7, 9, 11, 13, 15},
    ytick={1,3,5,7,9,11,13},
    yticklabels={0.14,0.12,0.10,0.08,0.06,0.04,0.02},
]

\addplot [
    matrix plot*,
    mesh/cols=15,
    point meta=explicit,
] table [
    x=x,
    y=y,
    meta=z
] {datat/heatmap_triples_updated2.dat};

\end{axis}
\end{tikzpicture}
\caption{Stopping time as a function of maximum allowed slosh and rod length.}
\label{fig:heatmap_stopping_time}
\end{figure}
The results are depicted in Figure \ref{fig:heatmap_stopping_time}. The figure shows that stopping time decreased steadily as the allowed sloshing angle, $\beta_{\max}$, increased. The fastest stopping times were achieved for rod lengths 0.05-0.07 m and $\beta_{\max} = 15^\circ$. Violations of the no-spill constraint $\beta < \beta_{\max}$ remained small, with the maximum violation averaging $0.22^\circ$. For the most restrictive case, $\beta_{\max} = 1^\circ$, the average maximum violation was $0.55^\circ$. This suggests that, for the considered trajectory, keeping the maximum slosh within very restrictive limits was challenging even though the controller was allowed to increase the stopping time freely. For this trajectory, smaller rod lengths ($l \leq 0.01$ m) required penalizing $\dot{\theta}_p$ and $\dot{\phi}_p$ in the cost function, as described in Section~\ref{sec:qp_formulation}. Because this penalization affects the stopping time, these cases were excluded from the analysis.

The second simulation experiment evaluated the robustness of the proposed method to errors in rod-length estimation. Such errors may arise in practice from uncertainty in the measured container geometry or from discrepancies between the real container shape and its idealized model. The true rod length was set to 50 mm, the maximum allowed sloshing angle to $5^\circ$, and the rod length used by the model was intentionally misspecified by 10\%--50\% in both the positive and negative directions. As a result, the sloshing angle violation increased approximately linearly with the modeling error (an error of $x$ \% caused a violation of $x$ \%) and was nearly symmetric with respect to the error direction. Even for the worst case of 50 \% error, the maximum violation remained about $2.4^\circ$, indicating that the method is relatively robust to rod-length uncertainty.

We further evaluated robustness to violations of the initial assumption that the liquid surface is perpendicular to the container's vertical axis. Initial offsets of $\beta_0 = 1^\circ \text{ to }5^\circ$ were imposed, and the emergency stop was simulated under the same trajectory and conditions as above. When the offset was created so that $\theta_{\text{offset}} = -\phi_{\text{offset}}$, the sloshing angle violation behaved linearly to $\beta_0$. When the offset was created so that $\theta_{\text{offset}} = \phi_{\text{offset}}$, the sloshing angle violation remained below $1^\circ$ for initial offsets up to $\beta_0 = 3.3^\circ$ and reached $2.6^\circ$ in the most severe case, $\beta_0 = 5^\circ$. These results indicate that the proposed method is not particularly sensitive to initial surface-orientation errors.

Finally, we evaluated a variety of pre-stop motion trajectories to generate a diverse set of initial conditions for the model. In total, 10 trajectories were tested, differing in acceleration profile, direction, and magnitude, resulting in linear and angular speeds up to 1.63 m/s and 0.83 rad/s at the stopping point. We performed the tests with $\beta_{\max}=5^\circ$ and $\beta_{\max} = 3^\circ$. With $\beta_{\max} = 5^\circ$, the highest maximum slosh constraint violation was $0.33^\circ$. With $\beta_{\max} = 3^\circ$, the most aggressive trajectory ($\mathbf{v} = [1.04, -0.91, 0.87, 0.12, 0.12, 0]$~[m/s;rad/s] at the stopping point) resulted in a $0.61^\circ$ constraint violation. The second-highest maximum violation was $0.32^\circ$. The stopping time was influenced primarily by the initial velocity and the permitted maximum sloshing angle, although the initial container angle also appeared to have an effect. The results verify that the proposed method does generalize to diverse initial conditions. For comparison, if the liquid modeling and no-spill constraints are removed and the method simply minimizes the cumulative velocity and jerk cost, the simulated maximum slosh reaches several tens of degrees for aggressive trajectories, reaching a region where the underlying simulation model no longer matches reality.

\subsection{Real-world experiments}
The real-world experiments aimed to validate the proposed method on real hardware and liquid. We conducted the experiments using a 7-DoF Franka Emika Panda robot. Accordingly, we set the joint space kinematic limits in (\ref{eq:joint_space_kin1})--(\ref{eq:joint_space_kin3}) as specified in the Franka hardware specification \cite{franka_robot_specs}. We use the same Cartesian constraints as in Section \ref{sec:simulation}. For the liquid, we used a cylindrical container with an inner diameter of 80 mm and filled it with water to heights of 30 mm, 60 mm, and 100 mm. We marked the original surface height on the outer edge of the container with a red marking to facilitate later analysis of the slosh angles. We recorded the experiments with a static video camera at 60 FPS and measured the maximum slosh angles from the footage. For each experiment, the camera was carefully aligned through practice runs so that the maximum slosh could be captured in clear close-up footage. We set $\beta_{\max} = 5^\circ$ for all experiments.  

We compared the proposed method with a baseline that omits slosh modeling and simply minimizes the cumulative velocity while using the same jerk cost. Keeping the jerk cost identical ensures that any differences are attributable to slosh modeling rather than to acceleration smoothing alone. Furthermore, earlier works on slosh-free and spill-free liquid manipulation do not provide a suitable baseline, as they rely on a predefined trajectory or goal pose and cannot be used to stop optimally from an arbitrary initial state when the goal is unknown a priori.

In these experiments, we used the Lissajous trajectory from~\cite{arrizabalaga2024geometric} as an example motion and performed emergency stops at two different points of the trajectory, creating varying initial conditions for testing. Specifically, we performed the stops at stopping point~1 with $\mathbf{v} = [-0.12, 0.32, 0.35, 0.35, 0.06, -0.01]$ (m/s and rad/s) and stopping point~2 with $\mathbf{v} =[-0.15, -0.06, -0.39, -0.31, 0.04, -0.06]$. We also ran the same trajectory with the same stopping points in simulation for comparison.

\begin{figure}
    \centering
    
    \begin{subfigure}[!ht]{0.4\textwidth}
        \centering
        \includegraphics[width=0.9\linewidth]{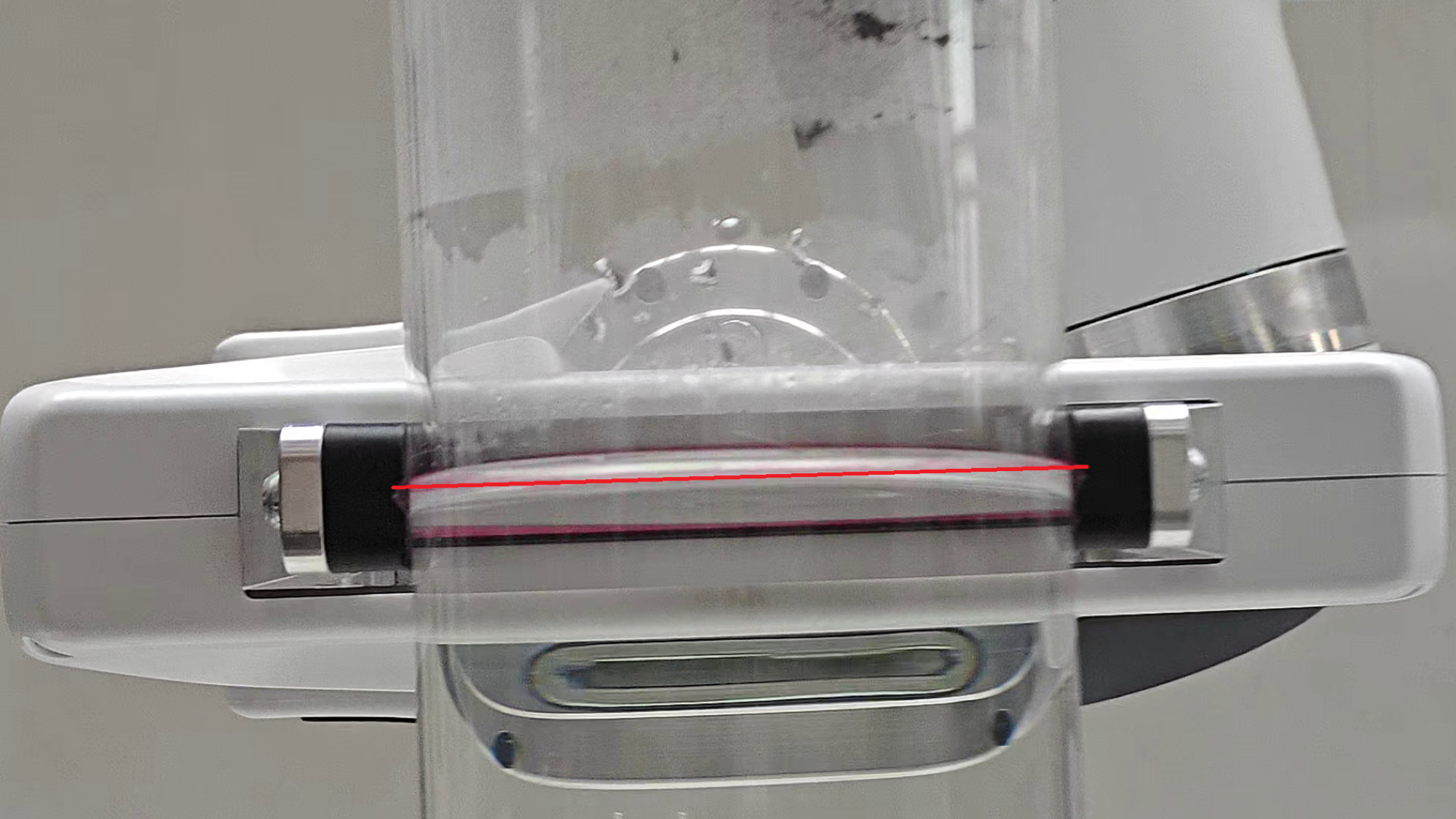}
        \captionsetup{justification=raggedright}
        \caption{Ours. The measured maximum slosh is $5.0^\circ$.}
        \label{fig:slosh_image_estop_exact}
    \end{subfigure}
    \hfill
    \begin{subfigure}[!ht]{0.4\textwidth}
        \centering
        \includegraphics[width=0.9\linewidth]{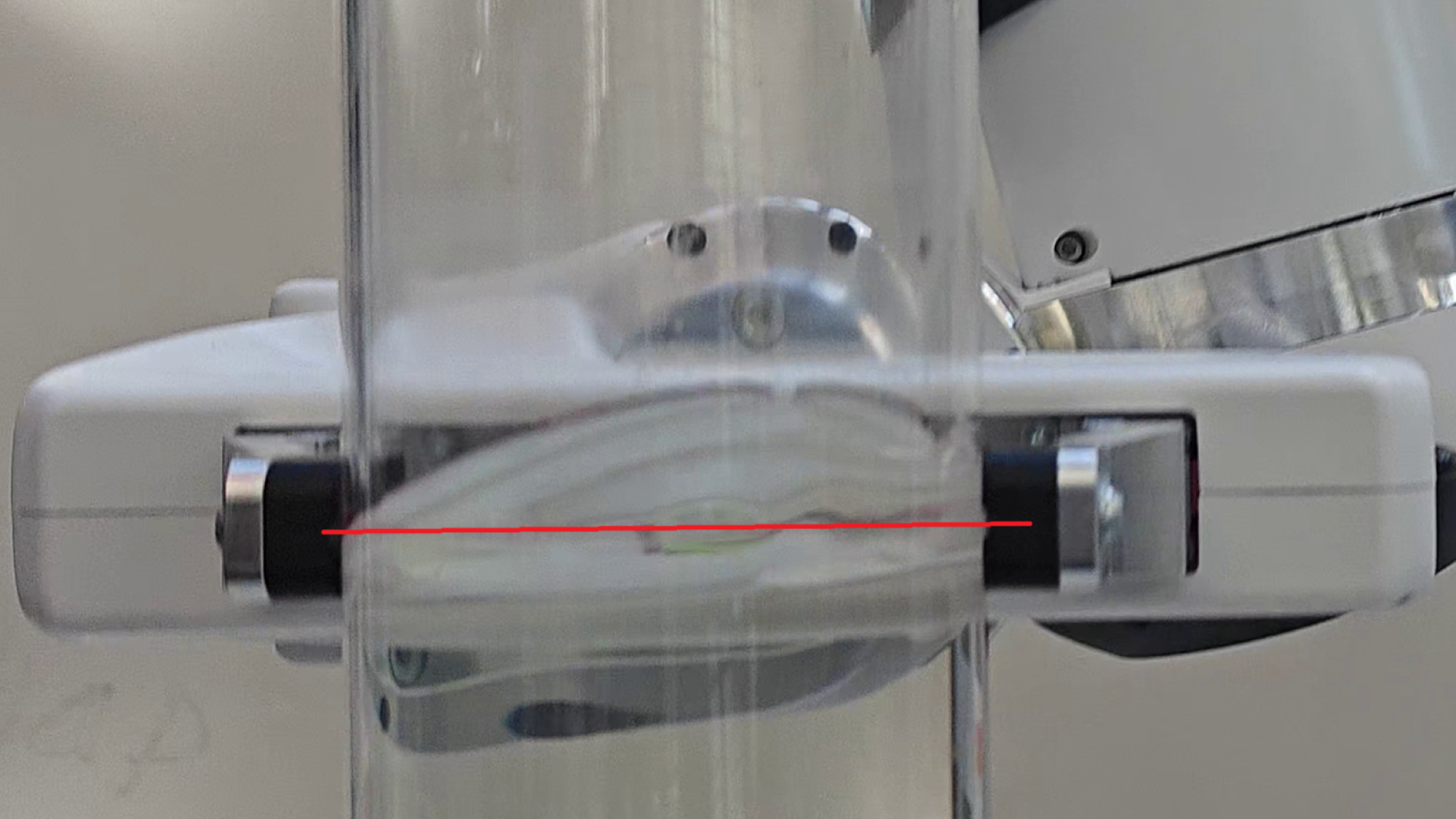}
        \captionsetup{justification=raggedright}
        \caption{Baseline. The measured maximum slosh is $26.4^\circ$.}
        \label{fig:slosh_image_baseline}
    \end{subfigure}
    \hfill
    \caption{Maximum sloshes for the proposed method and baseline. Stopping point 1, $h=60$ mm.}
    \label{fig:exact_vs_baseline}
\end{figure}

Table \ref{exp_results} summarizes the quantitative results, while Figure~\ref{fig:exact_vs_baseline} depicts a qualitative comparison of the maximum slosh between the proposed method and the baseline.

\begin{table*}[t]
\centering
\renewcommand{\arraystretch}{1.25}
\footnotesize
\setlength{\tabcolsep}{3.5pt}
\begin{tabular}{l|c|c|c|c|c|c|c|c|}
\cline{2-9}
\multicolumn{1}{c|}{}
& \multicolumn{4}{c|}{\textbf{Stopping point 1}}
& \multicolumn{4}{c|}{\textbf{Stopping point 2}} \\ \cline{2-9}

\multicolumn{1}{c|}{}
& \multicolumn{2}{c|}{\textbf{Ours}}
& \multicolumn{2}{c|}{\textbf{Baseline}}
& \multicolumn{2}{c|}{\textbf{Ours}}
& \multicolumn{2}{c|}{\textbf{Baseline}} \\ \cline{2-9}

\multicolumn{1}{c|}{}
& \textbf{Stopping time}
& \textbf{Slosh max}
& \textbf{Stopping time}
& \textbf{Slosh max}
& \textbf{Stopping time}
& \textbf{Slosh max}
& \textbf{Stopping time}
& \textbf{Slosh max} \\ \hline

\multicolumn{1}{|l|}{$h = 30$ mm}
& $1.02$ s & $5.9^\circ$ & $0.66$ s & $25.1^\circ$ & $0.69$ s&$6.2^\circ$  & $0.69$ s &  $9.6^\circ$\\ \hline

\multicolumn{1}{|l|}{$h = 60$ mm}
&  $0.99$ s& $5.0^\circ$ & $0.72$ s& $26.4^\circ$ & $0.68$ s& $5.7^\circ$ & $0.70$ s& $8.9^\circ$ \\ \hline

\multicolumn{1}{|l|}{$h = 100$ mm}
& $0.98$ s& $5.3^\circ$ & $0.72$ s& $21.3^\circ$& $0.70$ s& $5.0^\circ$& $0.69$ s & $8.3^\circ$ \\ \hline

\multicolumn{1}{|l|}{$h = 100$ mm (+50\% error)}
& $0.92$ s& $7.7^\circ$ & - & - & - & - & - & - \\ \hline

\multicolumn{1}{|l|}{$h = 100$ mm (-50\% error)}
& $1.07$ s& $3.9^\circ$ & - & - & - & - & - & - \\ \hline

\multicolumn{1}{|l|}{Simulation}
& $0.87$ s& $4.8^\circ$ & - & - & $0.70$ s& $4.4^\circ$& - & - \\ \hline

\end{tabular}

\caption{Results from the experiments with the Lissajous trajectory. Allowed slosh was 5 degrees, and the velocity tolerance for the stopping time measurements was $10^{-2}$ m/s. The rod length in the simulation was set so that it matches the real-world container with $h = 100$ mm.}
\label{exp_results}
\end{table*}
The figure and table show that the proposed method maintained the maximum sloshing angle close to the specified limit of $\beta_{\max}=5^\circ$. In contrast, the baseline method substantially exceeded the limit, producing maximum sloshing angles ranging from $8.3^\circ \text{ to }27.4^\circ$, depending on the stopping point and liquid fill height. For the proposed method (without intentional modeling error), only one case exceeded the limit by more than $1^\circ$. At stopping point 2 with $h=30$ mm, the measured violation was $1.2^\circ$. Further, the largest violation at stopping point 1 was also observed with $h=30$ mm, the violation being $0.9^\circ$. The results imply that the proposed model may slightly underestimate the real-world slosh when the ratio of liquid height to container radius is small, leading to overly aggressive stopping. 

At stopping point 1, the proposed method produced significantly slower stopping trajectories compared to the baseline. However, while stopping was approximately 40 \% slower, the method reduced the maximum sloshing angle by up to 71 \%. At stopping point 2, the proposed method was as fast as the baseline while still keeping sloshing within or close to the $5^\circ$ limit, unlike the baseline did. These results suggest that the proposed method did not merely smooth the deceleration profile compared to the baseline, but also utilized aggressive control actions when permitted by the predicted slosh dynamics. Measured velocity profiles during stopping for both methods and both stopping points are illustrated in Figure \ref{fig:velocity-comparison}.

\begin{figure}[!t]
\centering

\subfloat[Stopping point 1.\label{fig:vel-a}]{
\begin{minipage}{\columnwidth}
\centering
\begin{tikzpicture}
\begin{axis}[
    width=0.9\linewidth,
    height=0.55\linewidth,
    xlabel={Time (s)},
    ylabel={Velocity},
    grid=both,
    legend pos=north east,
    legend style={font=\scriptsize},
    legend image post style={scale=0.5},
    legend columns=2,
    no markers,
    xmin=0,
    xmax=1.5,
]

\addplot[red, thick] table [
    col sep=space,
    x expr=(\coordindex-500)/750,
    y index=0,
    each nth point=1,
    restrict x to domain=0:1.5,
    unbounded coords=discard,
] {datat/sp1_ours_baseline_vel2_combined.txt};
\addlegendentry{Ours $\dot{x}$}

\addplot[red, dashed, thick] table [
    col sep=space,
    x expr=(\coordindex-500)/750,
    y index=3,
    each nth point=1,
    restrict x to domain=0:1.5,
    unbounded coords=discard,
] {datat/sp1_ours_baseline_vel2_combined.txt};
\addlegendentry{Baseline $\dot{x}$}

\addplot[blue, thick] table [
    col sep=space,
    x expr=(\coordindex-500)/750,
    y index=1,
    each nth point=1,
    restrict x to domain=0:1.5,
    unbounded coords=discard,
] {datat/sp1_ours_baseline_vel2_combined.txt};
\addlegendentry{Ours $\dot{y}$}

\addplot[blue, dashed, thick] table [
    col sep=space,
    x expr=(\coordindex-500)/750,
    y index=4,
    each nth point=1,
    restrict x to domain=0:1.5,
    unbounded coords=discard,
] {datat/sp1_ours_baseline_vel2_combined.txt};
\addlegendentry{Baseline $\dot{y}$}

\addplot[black, thick] table [
    col sep=space,
    x expr=(\coordindex-500)/750,
    y index=2,
    each nth point=1,
    restrict x to domain=0:1.5,
    unbounded coords=discard,
] {datat/sp1_ours_baseline_vel2_combined.txt};
\addlegendentry{Ours $\dot{z}$}


\addplot[black, dashed, thick] table [
    col sep=space,
    x expr=(\coordindex-500)/750,
    y index=5,
    each nth point=1,
    restrict x to domain=0:1.5,
    unbounded coords=discard,
] {datat/sp1_ours_baseline_vel2_combined.txt};
\addlegendentry{Baseline $\dot{z}$}

\end{axis}
\end{tikzpicture}
\end{minipage}
}
\hfill
\subfloat[Stopping point 2.\label{fig:vel-b}]{
\begin{minipage}{\columnwidth}
\centering
\begin{tikzpicture}
\begin{axis}[
    width=0.9\linewidth,
    height=0.55\linewidth,
    xlabel={Time (s)},
    ylabel={Velocity},
    grid=both,
    legend pos=south east,
    legend style={font=\scriptsize},
    legend image post style={scale=0.5},
    legend columns=2,
    no markers,
    xmin=0,
    xmax=0.8,
]

\addplot[red, thick] table [
    col sep=space,
    x expr=(\coordindex-1000)/750,
    y index=0,
    each nth point=2,
    restrict x to domain=0:0.8,
    unbounded coords=discard,
] {datat/sp2_ours_baseline_vel_combined.txt}; 
\addlegendentry{Ours $\dot{x}$}

\addplot[red, thick, dashed] table [
    col sep=space,
    x expr=(\coordindex-1000)/750,
    y index=3,
    each nth point=2,
    restrict x to domain=0:0.8,
    unbounded coords=discard,
] {datat/sp2_ours_baseline_vel_combined.txt}; 
\addlegendentry{Baseline $\dot{x}$}

\addplot[blue, thick] table [
    col sep=space,
    x expr=(\coordindex-1000)/750,
    y index=1,
    each nth point=2,
    restrict x to domain=0:0.8,
    unbounded coords=discard,
] {datat/sp2_ours_baseline_vel_combined.txt}; 
\addlegendentry{Ours $\dot{y}$}

\addplot[blue, thick, dashed] table [
    col sep=space,
    x expr=(\coordindex-1000)/750,
    y index=4,
    each nth point=2,
    restrict x to domain=0:0.8,
    unbounded coords=discard,
] {datat/sp2_ours_baseline_vel_combined.txt}; 
\addlegendentry{Baseline $\dot{y}$}

\addplot[black, thick] table [
    col sep=space,
    x expr=(\coordindex-1000)/750,
    y index=2,
    each nth point=2,
    restrict x to domain=0:0.8,
    unbounded coords=discard,
] {datat/sp2_ours_baseline_vel_combined.txt}; 
\addlegendentry{Ours $\dot{z}$}

\addplot[black, thick, dashed] table [
    col sep=space,
    x expr=(\coordindex-1000)/750,
    y index=5,
    each nth point=2,
    restrict x to domain=0:0.8,
    unbounded coords=discard,
] {datat/sp2_ours_baseline_vel_combined.txt}; 
\addlegendentry{Baseline $\dot{z}$}

\end{axis}
\end{tikzpicture}
\end{minipage}
}

\caption{Measured linear velocity trajectories for emergency stops at both stopping points with $h=60$ mm. The velocities are measured by reading the robot joint velocities in the high-frequency control loop and converting them to Cartesian space.}
\label{fig:velocity-comparison}
\end{figure}
Similarly to the simulation experiments, we tested the method's robustness to modeling errors in the real-world by deliberately over- and under-specifying the rod length by 50 \%. We ran the experiments with $h = 100$ mm at stopping point 1. As a result, a rod length set too large caused overly aggressive braking that led to a slosh constraint violation. The measured maximum slosh was $7.7^\circ$, exceeding the $5^\circ$ limit by 54 \%. When the rod length was set 50 \% too short, the model produced excessively cautious braking, utilizing only $3.9^\circ$ of the available $5^\circ$ sloshing limit. This behavior is also reflected in the stopping times. The experiment with positive error produced 6.1 \% shorter stopping time compared to the case without modeling error, while the negative error increased the stopping time by 9.2 \%.

To summarize, the real-world experiments validated that the proposed method works with real robot hardware and real liquid. Only for the smallest liquid height, $h = 30$ mm, did the method appear to slightly underestimate the real-world sloshing, resulting in slosh constraint violations of approximately $1^\circ$. In contrast, the jerk-penalized baseline produced overly aggressive braking profiles, leading to large slosh constraint violations. Video recordings of example runs from the real-world experiments are provided in the supplementary material.

\section{Conclusions}
\label{sec:conclusions}
We proposed an emergency stop system for robots that manipulate liquids in open containers. The proposed method brings the robot to a stop in a time-optimal manner upon an emergency stop trigger, while ensuring that no liquid is spilled, as validated through both simulation and real-world experiments. By formulating the task as a quadratic optimization problem, the method runs in real time and enables an instant reaction when needed. 

The proposed method assumes that the robot is free to move in any direction during stopping and that all directions are equally hazardous. In some applications, the robot workspace may be constrained, limiting its ability to select optimal braking actions. Moreover, hazards can be directional, such that motions along certain directions should be penalized more than others. Further research is needed to expand the current work in this direction. 

Another limitation of the proposed method is that it implicitly assumes the containers to be filled close to the brim, keeping the allowed slosh within the range where the small-angle approximation of the pendulum model is valid. In addition, the spherical pendulum model can be analytically fitted only to containers with simple geometries, such as cylinders or rectangular cuboids. However, real-world containers, such as coffee or beer mugs, often have more complex geometries. This complicates parameter tuning, making it largely experimental or even rendering the model inapplicable. To date, complex container geometries have received little attention in the literature, and future work is needed to understand their effect on both planning of slosh-free trajectories as well as emergency stop behaviors. An interesting avenue for exploration is the role of sensing and feedback in this context.


\section*{Acknowledgements}
The authors acknowledge the use of MIDAS infrastructure of Aalto School of Electrical Engineering. V. Kyrki acknowledges the research environment provided by ELLIS Institute Finland.
\bibliographystyle{IEEEtran}
\bibliography{refs}

\end{document}